\documentclass[10pt,twocolumn,letterpaper]{article}

\usepackage{cvpr}
\usepackage{times}
\usepackage{epsfig}
\usepackage{graphicx}
\usepackage{amsmath}
\usepackage{amssymb}

\usepackage[table]{xcolor}
\usepackage{colortbl}
\usepackage{graphicx}
\usepackage{booktabs}
\usepackage{array}
\usepackage{caption}
\usepackage{tabularx}
\usepackage{multirow}
\usepackage{makecell}
\usepackage{epstopdf}
\usepackage{subcaption}
\usepackage{diagbox}

\definecolor{lightpink}{rgb}{1.0, 0.7, 0.7}
\definecolor{mistyrose}{rgb}{1.0, 0.89, 0.88}
\definecolor{lightyellow}{rgb}{1.0, 1.0, 0.8}
\definecolor{lightcyan}{rgb}{0.8, 1.0, 1.0}
\definecolor{mediumturquoise}{rgb}{0.28, 0.82, 0.8}

\usepackage{xcolor}
\definecolor{highlightred}{HTML}{e41a1c}
\definecolor{highlightorange}{HTML}{ff7f00} 
\definecolor{highlightyellow}{HTML}{ffff33} 
\definecolor{highlightgreen}{HTML}{4daf4a}
\definecolor{highlightper}{HTML}{984ea3}
\definecolor{highlightblue}{HTML}{377eb8}
\definecolor{highlightbrown}{HTML}{bf5b17}

\begin{document}

\title{A Comprehensive Re-Evaluation of Biometric Modality Properties \\in the Modern Era}

\author{Rouqaiah Al-Refai, Pankaja Priya Ramasamy, Ragini Ramesh, Patricia Arias-Cabarcos\thanks{Current affiliation: European Commission, Joint Research Centre (JRC), Ispra, Italy.}, Philipp Terh\"{o}rst \\
Paderborn University, Germany \\
{\tt\small Corresponding Author: rouqaiah.al.refai@uni-paderborn.de}
}

\maketitle
\thispagestyle{empty}

\begin{abstract}
\vspace{-0.5em}
The rapid advancement of authentication systems and their increasing reliance on biometrics for faster and more accurate user verification experience, highlight the critical need for a reliable framework to evaluate the suitability of biometric modalities for specific applications. Currently, the most widely known evaluation framework is a comparative table from 1998, which no longer adequately captures recent technological developments or emerging vulnerabilities in biometric systems. To address these challenges, this work revisits the evaluation of biometric modalities through an expert survey involving 24 biometric specialists. The findings indicate substantial shifts in property ratings across modalities. For example, face recognition, shows improved ratings due to technological progress, while fingerprint, shows decreased reliability because of emerging vulnerabilities and attacks. Further analysis of expert agreement levels across rated properties highlighted the consistency of the provided evaluations and ensured the reliability of the ratings. Finally, expert assessments are compared with dataset-level uncertainty across 55 biometric datasets, revealing strong alignment in most modalities and underscoring the importance of integrating empirical evidence with expert insight. Moreover, the identified expert disagreements reveal key open challenges and help guide future research toward resolving them. 
\vspace{-0.9em}
\end{abstract}

\section{Introduction}

\begin{table}[!t]
    \centering
    \caption{\textbf{Re-evaluation of Properties of Biometric Modalities}
    Shown is the (a) qualitative assessments from Jain et al. (1998) \cite{jain1999biometrics} based on 3 experts and (b) our updated (2025) quantitative assessment based on 24 biometric experts.
    Ratings range from \colorbox{lightpink}{Very Low} (1.0-1.8), \colorbox{mistyrose}{Low} (1.9-2.6), \colorbox{lightyellow}{Medium} (2.7-3.4), \colorbox{lightcyan}{High} (3.5-4.2), and \colorbox{mediumturquoise}{Very High} (4.3-5.00). The updated ratings reveal significant improvements in most modalities.} 
    \label{tab:past_present}
    \renewcommand{\arraystretch}{0.8}
    \setlength{\tabcolsep}{1.0 pt}  
    \footnotesize
    \begin{subtable}[t]{0.49\textwidth}
        \centering
        \caption{Qualitative assessment from Jain et al. (1998) based on 3 experts}
        \label{tab:1}
        \begin{tabular}{lccccccc}
            \Xhline{0.8pt}
            \diagbox[width=2.0cm, height=1.0cm, innerleftsep=1pt, innerrightsep=0.5pt]{Modality}{Properties} & 
            \makecell{\rotatebox{70}{Acceptability}} & 
            \makecell{\rotatebox{70}{Circumvention}} & 
            \makecell{\rotatebox{70}{Collectability}} & 
            \makecell{\rotatebox{70}{Performance}} & 
            \makecell{\rotatebox{70}{Permanence}} & 
            \makecell{\rotatebox{70}{Uniqueness}} & 
            \makecell{\rotatebox{70}{Universality}} \\ 
            \midrule
            DNA         & \cellcolor{mistyrose} L & \cellcolor{mistyrose} L & \cellcolor{mistyrose} L & \cellcolor{lightcyan} H & \cellcolor{lightcyan} H & \cellcolor{lightcyan} H & \cellcolor{lightcyan} H \\
            Ear         & \cellcolor{lightcyan} H & \cellcolor{lightyellow} M & \cellcolor{lightyellow} M & \cellcolor{lightyellow} M & \cellcolor{lightcyan} H & \cellcolor{lightyellow} M & \cellcolor{lightyellow} M \\
            Face        & \cellcolor{lightcyan} H & \cellcolor{mistyrose} L & \cellcolor{lightcyan} H & \cellcolor{mistyrose} L & \cellcolor{lightyellow} M & \cellcolor{mistyrose} L & \cellcolor{lightcyan} H \\
            Fingerprint & \cellcolor{lightyellow} M & \cellcolor{lightcyan} H & \cellcolor{lightyellow} M & \cellcolor{lightcyan} H & \cellcolor{lightcyan} H & \cellcolor{lightcyan} H & \cellcolor{lightyellow} M \\
            Gait        & \cellcolor{lightcyan} H & \cellcolor{lightyellow} M & \cellcolor{lightcyan} H & \cellcolor{mistyrose} L & \cellcolor{mistyrose} L & \cellcolor{mistyrose} L & \cellcolor{lightyellow} M \\
            Hand Geometry    & \cellcolor{lightyellow} M & \cellcolor{lightyellow} M & \cellcolor{lightcyan} H & \cellcolor{lightyellow} M & \cellcolor{lightyellow} M & \cellcolor{lightyellow} M & \cellcolor{lightyellow} M \\
            Hand Vein   & \cellcolor{lightyellow} M & \cellcolor{lightcyan} H & \cellcolor{lightyellow} M & \cellcolor{lightyellow} M & \cellcolor{lightyellow} M & \cellcolor{lightyellow} M & \cellcolor{lightyellow} M \\
            Iris        & \cellcolor{mistyrose} L & \cellcolor{lightcyan} H & \cellcolor{lightyellow} M & \cellcolor{lightcyan} H & \cellcolor{lightcyan} H & \cellcolor{lightcyan} H & \cellcolor{lightcyan} H \\
            Keystroke   & \cellcolor{lightyellow} M & \cellcolor{lightyellow} M & \cellcolor{lightyellow} M & \cellcolor{mistyrose} L & \cellcolor{mistyrose} L & \cellcolor{mistyrose} L & \cellcolor{mistyrose} L \\
            Retinal Scan   & \cellcolor{mistyrose} L & \cellcolor{mistyrose} L & \cellcolor{mistyrose} L & \cellcolor{lightcyan} H & \cellcolor{lightyellow} M & \cellcolor{lightcyan} H & \cellcolor{lightcyan} H \\
            Signature   & \cellcolor{lightcyan} H & \cellcolor{mistyrose} L & \cellcolor{lightcyan} H & \cellcolor{mistyrose} L & \cellcolor{mistyrose} L & \cellcolor{mistyrose} L & \cellcolor{mistyrose} L \\
            Voice       & \cellcolor{lightcyan} H & \cellcolor{mistyrose} L & \cellcolor{lightyellow} M & \cellcolor{mistyrose} L & \cellcolor{mistyrose} L & \cellcolor{mistyrose} L & \cellcolor{lightyellow} M \\ \Xhline{0.8pt} \\
        \end{tabular}
     \end{subtable}
     \hfill
     \begin{subtable}[t]{0.49\textwidth}
     \centering
    \caption{Our quantitative assessment (2025) based on 24 experts}
    \label{tab:2}
    \begin{tabular}{lccccccc}
        \Xhline{0.8pt}
        \diagbox[width=2.0cm, height=1.0cm, innerleftsep=1pt, innerrightsep=0.5pt]{Modality}{Properties} & 
        \makecell{\rotatebox{70}{Acceptability}} & 
        \makecell{\rotatebox{70}{Circumvention}} & 
        \makecell{\rotatebox{70}{Collectability}} & 
        \makecell{\rotatebox{70}{Performance}} & 
        \makecell{\rotatebox{70}{Permanence}} & 
        \makecell{\rotatebox{70}{Uniqueness}} & 
        \makecell{\rotatebox{70}{Universality}} \\
        \midrule
        DNA & \cellcolor{mistyrose}2.2 & \cellcolor{lightcyan}3.8 & \cellcolor{lightyellow}2.8 & \cellcolor{mediumturquoise}4.3 & \cellcolor{mediumturquoise}5.0 & \cellcolor{mediumturquoise}4.7 & \cellcolor{mediumturquoise}5.0 \\
        ECG & \cellcolor{lightyellow}3.0 & \cellcolor{lightyellow}2.7 & \cellcolor{mistyrose}2.3 & \cellcolor{lightyellow}2.7 & \cellcolor{lightyellow}2.7 & \cellcolor{lightcyan}3.7 & \cellcolor{mediumturquoise}5.0 \\
        EEG & \cellcolor{lightyellow}3.2 & \cellcolor{lightcyan}3.8 & \cellcolor{mistyrose}2.2 & \cellcolor{lightyellow}3.0 & \cellcolor{mistyrose}2.6 & \cellcolor{lightcyan}3.6 & \cellcolor{mediumturquoise}4.8 \\
        Ear & \cellcolor{lightcyan}3.6 & \cellcolor{lightyellow}3.1 & \cellcolor{lightcyan}3.7 & \cellcolor{lightyellow}3.3 & \cellcolor{lightyellow}3.4 & \cellcolor{lightcyan}3.9 & \cellcolor{mediumturquoise}4.6 \\
        Face & \cellcolor{lightcyan}4.0 & \cellcolor{mistyrose}2.5 & \cellcolor{mediumturquoise}4.8 & \cellcolor{lightcyan}4.2 & \cellcolor{lightyellow}3.4 & \cellcolor{lightcyan}4.0 & \cellcolor{mediumturquoise}5.0 \\
        Fingerprint & \cellcolor{lightcyan}4.2 & \cellcolor{lightyellow}3.0 & \cellcolor{lightcyan}3.9 & \cellcolor{mediumturquoise}4.5 & \cellcolor{mediumturquoise}4.3 & \cellcolor{mediumturquoise}4.7 & \cellcolor{mediumturquoise}4.4 \\
        Gait & \cellcolor{lightcyan}3.6 & \cellcolor{lightcyan}3.9 & \cellcolor{mediumturquoise}4.3 & \cellcolor{lightyellow}3.0 & \cellcolor{lightyellow}2.9 & \cellcolor{lightyellow}3.3 & \cellcolor{mediumturquoise}4.3 \\
        Hand Geometry & \cellcolor{lightcyan}3.5 & \cellcolor{lightyellow}2.8 & \cellcolor{mediumturquoise}4.5 & \cellcolor{lightyellow}3.0 & \cellcolor{lightcyan}3.8 & \cellcolor{lightcyan}3.8 & \cellcolor{mediumturquoise}4.5 \\
        Hand Vein & \cellcolor{lightcyan}3.5 & \cellcolor{lightcyan}4.0 & \cellcolor{lightyellow}3.2 & \cellcolor{lightcyan}4.0 & \cellcolor{mediumturquoise}4.3 & \cellcolor{mediumturquoise}4.5 & \cellcolor{mediumturquoise}4.8 \\
        Iris & \cellcolor{lightcyan}3.5 & \cellcolor{lightcyan}3.5 & \cellcolor{lightyellow}3.1 & \cellcolor{mediumturquoise}4.6 & \cellcolor{mediumturquoise}4.6 & \cellcolor{mediumturquoise}4.8 & \cellcolor{mediumturquoise}4.8 \\
        Keystroke & \cellcolor{lightcyan}4.2 & \cellcolor{mediumturquoise}4.8 & \cellcolor{lightcyan}4.2 & \cellcolor{lightcyan}4.0 & \cellcolor{lightcyan}3.8 & \cellcolor{mediumturquoise}4.5 & \cellcolor{mediumturquoise}4.5 \\
        Retinal Scan & \cellcolor{lightcyan}3.7 & \cellcolor{mediumturquoise}4.3 & \cellcolor{lightcyan}3.7 & \cellcolor{mediumturquoise}4.3 & \cellcolor{mediumturquoise}4.3 & \cellcolor{mediumturquoise}4.3 & \cellcolor{mediumturquoise}4.3 \\
        Signature & \cellcolor{mediumturquoise}4.6 & \cellcolor{lightyellow}2.8 & \cellcolor{mediumturquoise}4.4 & \cellcolor{lightcyan}4.0 & \cellcolor{lightcyan}3.6 & \cellcolor{lightcyan}4.2 & \cellcolor{lightcyan}4.2 \\
        Voice & \cellcolor{lightyellow}3.4 & \cellcolor{mistyrose}2.2 & \cellcolor{lightcyan}4.0 & \cellcolor{lightcyan}3.6 & \cellcolor{lightyellow}3.2 & \cellcolor{lightcyan}4.1 & \cellcolor{lightcyan}4.0 \\
        \Xhline{0.8pt} 
    \end{tabular}
    \end{subtable}
\end{table}
Biometric recognition is a widely adopted technology utilized mainly for identifying individuals based on their physiological or behavioral characteristics. It offers a reliable alternative to traditional authentication methods such as passwords and tokens \cite{rubab2011biometrics}. Significant improvements in the accuracy and processing speed of biometric systems have enabled their broad adoption in various high-security areas, including border control \cite{kwon2008biometric}, forensics \cite{fattahi2021damaged}, and financial transactions \cite{onesi2024fintech}.
However, the practical deployment of biometric technologies relies not only on minimizing error rates but also on a broader set of fundamental properties that define the effectiveness, usability, and security of these systems \cite{alrousan2020comparative}. According to the foundational framework on biometrics introduced by Jain et al. \cite{jain1999biometrics}, for a biological measurement to be considered biometric, it must satisfy several essential properties defined as:
\begin{itemize}
\setlength\itemsep{-0.3mm}
    \item \textbf{Universality}: The presence of a trait in all individuals. 
    \item \textbf{Uniqueness}: The distinctiveness of a trait between individuals.
    \item \textbf{Permanence}: The stability of a specific trait over time.
    \item \textbf{Collectability}: The capability of measuring a trait with reasonable effort and cost.
    \item \textbf{Performance}: The ability to achieve accurate and efficient recognition utilizing the trait.
    \item \textbf{Acceptability}: The social and cultural opinion on the use of the trait.
    \item \textbf{Circumvention}: The resistance of the trait to fraud and spoofing attacks. 
\end{itemize}

Based on these criteria, several physiological and behavioral traits have been explored as biometric modalities. Common ones such as face, fingerprint, iris, hand geometry, hand vein, signature, and voice are widely accepted and deployed \cite{jain2004introduction}. Others, including Deoxyribonucleic Acid (DNA), electroencephalogram (EEG), electrocardiogram (ECG), gait, and keystroke dynamics \cite{ross2019some} are still emerging and attracting increasing research interest due to their potential in specialized domains.

An early qualitative assessment of 12 biometric modalities against the seven key properties was presented in \cite{jain1999biometrics}, based on the opinions of three domain experts. This assessment, illustrated in  Table \ref{tab:1}, categorized modalities using a low-medium-high rating scale. While foundational, it reflected the technological capabilities and knowledge understanding of the late 1990s. Since then, advances in machine learning, sensors, and signal processing have significantly transformed the capabilities of biometric systems.
For example, face recognition, initially evaluated as moderate in performance, currently achieves near-human and super-human accuracy using deep neural networks \cite{DBLP:journals/pami/OToolePJAPA07, doi:10.1073/pnas.1721355115}. Similarly, keystroke dynamics and EEG biometrics have evolved from considered experimental to feasible options in specific application domains \cite{s22186929, app132011478, BARAKU202417}. 
These shifts necessitate a systematic re-evaluation of biometric modalities that goes beyond recognition performance alone. It is important to reassess how each modality aligns with the seven key properties in the context of current technological advancements and user expectations. To address this, we conducted a comprehensive evaluation based on input from 24 biometric experts. Our contributions are as follows: 
\begin{itemize}
    \item \textbf{Re-evaluating 14 biometric modalities} (DNA, ear, ECG, EEG, face, fingerprint, gait, hand geometry, hand vein, iris, keystroke, retinal scan, signature, and voice) across the seven biometric properties using expert survey data. To ensure comparability across modalities despite uneven domain expertise, we utilize a matrix factorization approach to align expert ratings by leveraging overlaps between modalities. 
    \item \textbf{Assessing the level of expert agreement} for each property-modality pair through variance analysis, providing insight into the consistency and reliability of expert opinions.
    \item \textbf{Investigating the alignment between expert perceptions and empirical evidence} of modality performance across 55 biometric datasets to identify discrepancies, validate expert judgments, and inform future biometric system design and evaluation.   
    \vspace{-0.5em}
\end{itemize}
This framework provides a timely update to Table \ref{tab:1} \cite{jain1999biometrics}, offering a more accurate and application-relevant understanding of biometric modality characteristics. Table \ref{tab:2} summarizes the updated expert assessments and can be directly compared with the original evaluations to observe changes across modalities. The updated ratings are intended to support both researchers and practitioners in selecting appropriate biometric systems for specific use cases.
\section{Related Work}
Research on biometric modality evaluation has traditionally centered around seven fundamental properties. The foundational work by Jain et al. \cite{jain1999biometrics} introduced a comparative framework assessing biometric traits based on universality, uniqueness, permanence, collectability, performance, acceptability, and circumvention. Their qualitative evaluation, summarized in Table \ref{tab:1}, offered early guidance for selecting appropriate modalities in biometric applications. Subsequent studies expanded this framework, by considering emerging modalities and technological advancements. For instance, the evaluation conducted in \cite{alrawili2024comprehensive} reported slight improvements in the universality and uniqueness of ear biometrics, as well as enhanced permanence and performance for signature-based recognition. Similarly, Harakannanavar et al.~\cite{harakannanavar2019comprehensive} introduced a comparison table with the addition of a measurability property and the study in \cite{sabhanayagam2018comprehensive} followed a similar approach, by comparing several modalities against the original seven properties. However, despite these contributions, these studies \cite{alrawili2024comprehensive, harakannanavar2019comprehensive, sabhanayagam2018comprehensive} lacked transparency in explaining the methodologies used for assigning ratings, raising concerns about their subjectivity and reproducibility. 

Other efforts expanded beyond the original framework by incorporating new evaluation factors such as cancellability and security \cite{vandehaar2013characteristics} or by focusing on reviewing recognition techniques and public datasets without in-depth analysis of modality properties. \cite{unar2014review}. More recent work by Gui et al. \cite{gui2019survey} explored emerging modalities like brainwave biometrics, highlighting their potential in high-security applications despite limited public acceptance. While these studies provided valuable insights into specific biometric traits, many restricted their evaluations to a subset of modalities or lacked systematic, side-by-side comparisons across all seven core properties.

In parallel to these developments, ensuring the reliability of biometric performance metrics has become increasingly important. Several studies have addressed the statistical reliability of such metrics, focusing on confidence interval estimation, sample size determination, and fairness testing across demographic groups \cite{mansfield2002best, ross2006confidence, bolle2004sample, han2014statistical, le2017elasticity}. The uncertainty analysis presented in \cite{fallahi2025reliability} introduced the BioQuake metric, which  quantifies the deviation between reported and actual error rates across biometric datasets. This analysis revealed significant inconsistencies across numerous biometric datasets, emphasizing the importance of complementing expert assessments with empirical evaluations grounded in robust statistical methods. 

While prior efforts offered important contributions, a significant gap remains: the absence of comprehensive, transparent, and systematic studies evaluating all seven biometric properties across a wide range of modalities in the age of high-performance AI. Addressing this gap, the paper presents an updated, expert-driven comparative analysis across 14 commonly researched modalities. By combining expert ratings with careful methodological design, the aim is to support more reliable evaluations of biometric systems.

\section{Method and Experimental Setup}
\subsection{Overview}
An expert survey was conducted to collect ratings for each biometric property based on the extent to which each property stands true for each modality. Experts were initially selected based on publication impact, including highly cited papers on specific modalities and overall citations on Google Scholar. We first contacted 200 researchers with over 200 citations, then added 143 more experts identified from program committees of top biometric venues (ICB, IJCB, BTAS, TBIOM). All participants were informed that responses would remain confidential and handled per the European data protection regulations. The survey was accessible only after providing consent via an ethics form.

\subsection{Expert Survey Design and Distribution}
\subsubsection{Survey Design}
The expert survey was designed and administered using Lime Survey, a secure web-based tool. The survey consists of two main components: quantitative and qualitative questions. The quantitative section consists of carefully designed questions to capture expert ratings based on their knowledge and expertise. The qualitative section is composed of open-ended questions to gain deeper insights into the experts' evaluations. Responses allow experts to provide reasoning for their assessments. This combination of quantitative ratings and qualitative feedback enables a more thorough justification and analysis of collected data.

The survey includes 47 questions, distributed consistently across the assessed biometric modalities. 
The complete list of survey questions is provided in the Appendix. The survey begins with a modality selection question (Q1), where experts select the biometric modalities they have experience with, as they are not required to evaluate all available modalities. The assessment covers 14 biometric modalities: DNA, Ear, ECG, EEG, Face, Fingerprint, Gait, Hand Geometry, Hand Vein, Iris, Keystroke, Retinal, Signature, and Voice. Following the modality selection, the main section of the survey (Q2-Q43) asks experts to rate seven biometric properties for each selected modality using a 5-point Likert scale. This scale was chosen to capture experts' opinions on how strongly each property applies to the modality. It also allows clear differentiation between modalities with subtle differences in properties, for instance, while both DNA and Ear have high universality, DNA is more universally present, which the scale helps capture. The 5-point scale ratings are defined as: 1 - Very Low, 2 - Low, 3 - Medium, 4 - High, and 5 - Very High. The questions were presented in a matrix format, with properties as rows and rating levels as columns. A 5-point scale was preferred over a 7-point scale to avoid overwhelming participants and random responses. The remaining questions in this section are qualitative, asking experts to explain their ratings and indicate their level of confidence. To support future research directions, the survey included two additional questions (Q44-Q45), inviting experts to suggest emerging biometric modalities that are not covered in this study and provide feedback on how the assessment methodology could be improved. Finally, to gain a basic understanding of the participant's background, the survey ends with two questions (Q46-Q47) about demographic information, especially age and region of residence.
\vspace{-0.5em}

\subsubsection{Pilot Testing and Survey Distribution}
Before launching the main survey, a pilot test with three experts was conducted to check the questions clarity, formatting, and estimate completion time. Feedback from the pilot test, resulted in minor revisions, including improvements to the open-ended reasoning section. Specifically, the interface was updated so the selected rating auto-filled the corresponding reasoning field, avoiding the need to switch back and forth.
The estimated time for completing the survey was approximately 10 minutes, and the exported responses from the survey matched the expected data structure.

For survey distribution, experts were identified through their publications and publicly available professional profiles, such as LinkedIn. Personalized email invitations were prepared using a word mailing tool and contact was initiated via email and LinkedIn messaging. Two weeks after the initial outreach, follow-up messages were sent again to encourage participation and increase the response rate.

\subsection{Data Preparation and Missing Value Handling}

The structure of the survey—where experts only rated modalities they were familiar with— resulted in missing values across the rating matrix, a table where each row represents an expert and each column corresponds to a specific modality-property pair (e.g., face-uniqueness). Managing these missing values is critical to ensure the robustness of subsequent analysis and interpretations. 
\vspace{-0.5em}

\subsubsection{Type of Missing Data}
In statistical analysis, understanding missing data is key to selecting the right handling strategy. According to \cite{little2002statistical, little2019statistical}, missing data falls into three categories: Missing Completely at Random (MCAR), Missing at Random (MAR), and Missing Not at Random (MNAR). Due to participants evaluating only specific biometric modalities within their expertise, the survey design inherently lead to missing values, as experts may skip rating modalities outside their domain. This qualifies the missing data as MNAR, where the likelihood of a value being missing depends on an unobserved variable, in this case the expert's level of expertise.

The survey output includes expert ratings for each modality-property combination. Initial analysis of the response distribution revealed a highly uneven spread of ratings across modalities, driven by experts evaluating only the modalities within their area of expertise. This reinforces the MNAR nature of the missing data, as the absence of a rating is not random but systematically tied to the respondent’s domain knowledge.
It was also noticeable that over 30\% of the total ratings were missing across modalities. Therefore, to ensure reliable and robust evaluation of survey outcomes under MNAR setting, the analysis was conducted in two ways. One way is through analyzing observed ratings only, where for each modality-property pair, the mean rating was computed by averaging across the experts who evaluated that specific modality. This keeps the missing values intact to avoid potential biases introduced through imputation. The other way is through using matrix factorization to align the ratings between the different modalities.
\vspace{-0.5em}

\subsubsection{Aligning Expert Ratings Across Modalities}
While missing values in expert ratings are manageable when analyzing individual modalities, they present a challenge when considering response sparsity. Since each expert rates only a subset of modalities, the raw rating matrix lacks a shared baseline: modality A may be rated by one group of experts, and modality B rated by another, with no direct link between them. Matrix factorization was employed to address this challenge by aligning these ratings across modalities using overlapping expert assessment.

Technically, matrix factorization is a dimensionality reduction technique used in collaborative filtering. Given a large, sparse expert rating matrix \( R \in \mathbb{R}^{m \times n} \), with \( m \) experts and \( n \) property-modality pairs, the goal is to find two lower-rank matrices \( M \in \mathbb{R}^{m \times r} \) and \( N \in \mathbb{R}^{n \times r} \) by solving the following optimization problem:
\[
\min_{M \geq 0,\, N \geq 0} \| R - M N^T \|_F^2
\] 
\noindent
Here, \( \| \cdot \|_F \) denotes the Frobenius norm, which measures reconstruction error. The approximated original matrix is:
\[
R \approx M \cdot N^{T}
\]
\noindent
The rank \( r \) is a hyperparameter that controls the dimensionality of the latent space, where \( M \) captures latent expert traits and \( N \) captures latent modality-property structures. Given that the expert ratings fall on a bounded, non-negative scale (1 to 5), we specifically employed Non-Negative Matrix Factorization (NMF) \cite{mccartney2022sparse, lee1999learning} to ensure interpretability and meaningful reconstruction. Before applying NMF, missing values were initialized using iterative imputation with Bayesian ridge regression, a method that iteratively predicts missing entries based on observed correlations between features. This initialization step is necessary to produce a fully numeric input matrix suitable for factorization.

\section{Results and Discussion}
\subsection{Expert Survey Response Statistics}

This section summarizes expert responses collected for the comparative analysis of biometric modalities. Of 343 invited experts, 24 submitted complete responses, yielding a 7\% response rate. Despite the low response rate, the responses quality was high, with most participants having experience in multiple biometric modalities, enabling a broad yet informative evaluation. Demographically, 75\% were mid-career professionals aged 31-60, indicating substantial domain experience. Geographically, 66.67\% of experts reside in Europe, 16.67\% in North America and 8.33\% in Asia, suggesting a largely international panel. 

\begin{figure}[t]
    \includegraphics[width=0.45\textwidth]{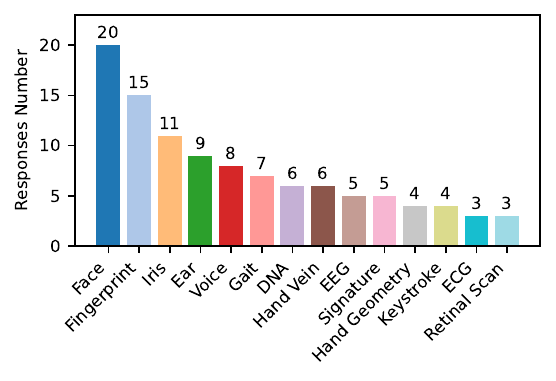}
    \caption{\textbf{Expert Responses Distribution Across Modalities -} 
    Face, fingerprint, and iris are the most evaluated, while behavioral and emerging traits, such as ECG, received lower responses.
    } 
    \label{fig:distribution}
\end{figure}

The response distribution across the 14 biometric modalities was notably uneven as shown in Figure \ref{fig:distribution}. Face recognition was the most represented, selected by 83.33\% of experts, followed by fingerprint (62.5\%) and iris (45.83\%). While ear, voice and gait were moderately represented at 37.5\%, 33.33\% and 29.17\%, respectively. Other behavioral and emerging modalities, such as ECG and retinal scans, were selected by only 12.5\% of experts. This distribution reflects the dominance of face, fingerprint and iris in both research and real-world deployments, while the moderate presence of ear, voice, gate and DNA modalities highlights their growing exposure and future potential. 

Analysis of response frequency showed that most experts evaluated multiple biometric modalities, highlighting the interdisciplinary nature of biometric research. A co-occurrence analysis revealed strong overlap among physiological modalities, particularly face, fingerprint and iris recognition, implying that expertise in one often extends to others within the same domain. In contrast, behavioral modalities exhibited minimal overlap, indicating more specialized, less commonly shared expertise. Experts also reported consistently high to very high confidence in their ratings, reinforcing the reliability and credibility of the data.

\subsection{Quantitative Expert Analysis}
\begin{table}[]
    \centering
    \caption{\textbf{Aligned Expert Ratings on the Properties of Biometric Modalities -} 
    Via matrix factorization the expert ratings are aligned between the different modalities.
    Ratings ranges from \colorbox{lightpink}{Very Low} (1.0-1.8), \colorbox{mistyrose}{Low} (1.9-2.6), \colorbox{lightyellow}{Medium} (2.7-3.4), \colorbox{lightcyan}{High} (3.5-4.2), to \colorbox{mediumturquoise}{Very High} (4.3-5.0);}.\\
    \label{tab:rating_max_factor}
    \renewcommand{\arraystretch}{0.8}
    \setlength{\tabcolsep}{1.0 pt}  
    \footnotesize
    \begin{tabular}{lccccccc}
        \Xhline{0.8pt}
        \diagbox[width=2.0cm, height=1.0cm, innerleftsep=1pt, innerrightsep=0.5pt]{Modality}{Properties} &
        \makecell{\rotatebox{70}{Acceptability}} & 
        \makecell{\rotatebox{70}{Circumvention}} & 
        \makecell{\rotatebox{70}{Collectability}} & 
        \makecell{\rotatebox{70}{Performance}} & 
        \makecell{\rotatebox{70}{Permanence}} & 
        \makecell{\rotatebox{70}{Uniqueness}} & 
        \makecell{\rotatebox{70}{Universality}} \\
        \midrule
        DNA & \cellcolor{lightpink}1.8 & \cellcolor{lightcyan}4.1 & \cellcolor{lightyellow}2.7 & \cellcolor{mediumturquoise}4.4 & \cellcolor{mediumturquoise}5.0 & \cellcolor{mediumturquoise}4.7 & \cellcolor{mediumturquoise}5.0 \\
        ECG & \cellcolor{lightyellow}2.9 & \cellcolor{mistyrose}2.0 & \cellcolor{mistyrose}2.5 & \cellcolor{mistyrose}2.4 & \cellcolor{mistyrose}2.4 & \cellcolor{lightcyan}3.8 & \cellcolor{mediumturquoise}5.0 \\
        EEG & \cellcolor{lightyellow}3.2 & \cellcolor{lightyellow}3.3 & \cellcolor{lightpink}1.5 & \cellcolor{lightyellow}2.9 & \cellcolor{lightyellow}2.7 & \cellcolor{lightyellow}3.4 & \cellcolor{mediumturquoise}4.9 \\
        Ear & \cellcolor{lightcyan}3.9 & \cellcolor{lightyellow}3.2 & \cellcolor{lightcyan}4.2 & \cellcolor{lightyellow}3.4 & \cellcolor{lightyellow}3.2 & \cellcolor{lightcyan}3.7 & \cellcolor{mediumturquoise}4.6 \\
        Face & \cellcolor{lightcyan}4.1 & \cellcolor{mistyrose}2.6 & \cellcolor{mediumturquoise}4.8 & \cellcolor{mediumturquoise}4.3 & \cellcolor{lightyellow}3.4 & \cellcolor{lightcyan}4.0 & \cellcolor{mediumturquoise}5.0 \\
        Fingerprint & \cellcolor{lightcyan}4.1 & \cellcolor{lightyellow}3.1 & \cellcolor{lightcyan}4.0 & \cellcolor{mediumturquoise}4.6 & \cellcolor{mediumturquoise}4.4 & \cellcolor{mediumturquoise}4.7 & \cellcolor{mediumturquoise}4.4 \\
        Gait & \cellcolor{lightcyan}3.5 & \cellcolor{lightcyan}3.8 & \cellcolor{lightcyan}4.0 & \cellcolor{lightyellow}2.7 & \cellcolor{lightyellow}2.8 & \cellcolor{lightyellow}2.8 & \cellcolor{lightcyan}4.0 \\
        Hand Geometry & \cellcolor{lightcyan}3.5 & \cellcolor{lightyellow}2.9 & \cellcolor{mediumturquoise}4.5 & \cellcolor{lightyellow}3.2 & \cellcolor{lightcyan}3.7 & \cellcolor{lightcyan}3.9 & \cellcolor{mediumturquoise}4.5 \\
        Hand Vein & \cellcolor{lightcyan}4.2 & \cellcolor{mediumturquoise}4.5 & \cellcolor{lightcyan}3.6 & \cellcolor{mediumturquoise}4.4 & \cellcolor{lightcyan}4.2 & \cellcolor{mediumturquoise}4.5 & \cellcolor{mediumturquoise}4.9 \\
        Iris & \cellcolor{lightcyan}3.5 & \cellcolor{lightcyan}3.6 & \cellcolor{lightyellow}2.9 & \cellcolor{mediumturquoise}4.6 & \cellcolor{mediumturquoise}4.6 & \cellcolor{mediumturquoise}4.7 & \cellcolor{mediumturquoise}4.8 \\
        Keystroke & \cellcolor{lightcyan}3.9 & \cellcolor{mediumturquoise}4.6 & \cellcolor{mediumturquoise}4.5 & \cellcolor{lightcyan}4.0 & \cellcolor{lightcyan}3.6 & \cellcolor{mediumturquoise}4.5 & \cellcolor{mediumturquoise}4.8 \\
        Retinal Scan & \cellcolor{lightcyan}3.5 & \cellcolor{lightcyan}4.1 & \cellcolor{lightcyan}3.6 & \cellcolor{mediumturquoise}4.4 & \cellcolor{mediumturquoise}4.4 & \cellcolor{mediumturquoise}4.4 & \cellcolor{mediumturquoise}4.4 \\
        Signature & \cellcolor{mediumturquoise}4.6 & \cellcolor{mistyrose}2.5 & \cellcolor{lightcyan}4.0 & \cellcolor{lightcyan}3.7 & \cellcolor{lightyellow}3.1 & \cellcolor{mediumturquoise}4.3 & \cellcolor{lightcyan}4.1 \\
        Voice & \cellcolor{lightyellow}3.1 & \cellcolor{mistyrose}2.1 & \cellcolor{lightyellow}3.3 & \cellcolor{lightcyan}3.7 & \cellcolor{lightcyan}3.7 & \cellcolor{mediumturquoise}4.4 & \cellcolor{lightcyan}4.0 \\
        \Xhline{0.8pt}
    \end{tabular}
\end{table}

To analyze the expert survey responses, the collected ratings were summarized into two constructed tables. Table \ref{tab:2} is generated based on the original expert ratings and Table \ref{tab:rating_max_factor} is based on a reconstructed rating matrix obtained through matrix factorization. The raw ratings table \ref{tab:2} presents the mean ratings for each biometric property calculated directly from the raw expert responses, without applying any imputation methods. Table \ref{tab:rating_max_factor} was derived using matrix factorization which was utilized to estimate missing values and produce a complete rating set. This method helps mitigate the potential biases introduced by uneven response distributions and highlight properties where experts ratings may have undervalued or overestimated it. For instance, the collectability of EEG shows a notably higher rating in the raw ratings (2.2) compared to the matrix factorization version (1.5), indicating an overestimation. 

Comparing both tables demonstrates a strong degree of overlap between properties ratings with only minor variation across certain modalities. This consistency highlights the robustness of the employed methods for handling collected data. Based on rating from both tables, it is observed that some modalities can be ideal candidates for almost all biometric applications such as keystroke, retinal scans and hand vein which exhibit high ratings across all properties. Similarly, face, fingerprint, signature and iris consistently show high scores across most properties, indicating their suitability for both security and convenience-based applications. However, emerging modalities like ECG and EEG exhibit medium to low ratings, reflecting their ongoing development and need for further research.  

These trends imply that recent technological advancements have enhanced expert perception and confidence in biometric modalities potential. Compared to earlier assessments, such as the one presented in Table \ref{tab:1} from 1998 \cite{jain1999biometrics}, the ratings in Table \ref{tab:2} and Table \ref{tab:rating_max_factor} express significant improvements across key biometric properties. The methodology adopted in this study addresses several limitations of past evaluations by involving a broader and more diverse group of experts, introducing a finer-grained rating scale for more specific assessments and carefully handling missing data to reduce potential biases. These enhancements offer a more comprehensive and updated reassessment of biometric modality suitability for modern applications. 

\subsection{Qualitative Modality-Specific Expert Analysis}
To complement the quantitative rating, open-ended expert responses were analyzed to provide deeper insights into the evaluation of each biometric modality. This qualitative analysis offers an understanding of the key factors influencing expert judgments across biometric properties, highlighting strengths, challenges, and trends critical for understanding modality reliability in applications:

\textbf{DNA} biometrics received high ratings for universality, uniqueness, performance, and permanence, consistent with prior studies. Experts highlighted DNA's inherent universality and stability over time, though uniqueness is slightly qualified in cases involving identical twins or closely related individuals. Performance remains consistent with past comparisons, dependent on sample quality, while views on collectability diverged, with some citing easy sample collection and others highlighting sequencing costs. Circumvention received low ratings, with experts agreeing that while attacks are possible, they remain technically challenging. 

\textbf{ECG} biometrics received high ratings for universality and uniqueness, similar to prior studies, while other properties showed some variation. Experts emphasized the ease of obtaining ECG signals from individuals, though uniqueness can be affected by similar heart conditions among individuals. Permanence and performance were rated low due to the effect of external influences such as stress and illnesses. Similarly, collectability is seen as challenging given the need for medical equipment for collecting data, and acceptability is rated medium due to the intrusive nature of the collection. Opinions over circumvention varied significantly with the provided rating, which reflects uncertainty and raises concerns about the reliability of these ratings.

\textbf{EEG} biometrics showed high ratings for universality, and medium ratings across most of the other modalities, aligning with prior research. Experts agree that EEG signals are present in all individuals, but distinguished uniqueness is dependent on extracted features. Permanence, rated as medium, is impacted by factors like aging and emotional state. Collectability is rated low due to the specialized equipment required for data collection. The performance remains debated, with concerns about noisy signals and generalizability. Acceptability is rated medium because of fears about mind-reading and privacy, although one expert noted that there is a growing public interest.

\textbf{Ear} biometrics received a consistently high rating category for acceptability, collectability, uniqueness, and universality, with slight variations in other properties. Experts highlighted that ear shape remains stable throughout most of the lifespan, though changes can occur in early life stages and old age. Collectability is rated high for external ear images but can be a challenge for internal data, which requires special equipment. The performance and circumvention are rated medium, where one is influenced by occlusions or image quality, and the other is prone to spoofing vulnerabilities. Although acceptability is rated as high, it received mixed views—some find it less intrusive than face biometrics, while others consider ear images unusual. 

\textbf{Face} biometrics exhibited high ratings across universality, uniqueness, collectability, accessibility, and performance, with slightly varied permanence and circumvention ratings. Experts emphasized that facial data is easily collected and widely available. However, uniqueness can be challenging in the case of identical twins and relatives. Since faces can be affected by aging and surgeries, the permanence is still rated as medium, similar to prior works. With the widespread use of smartphones and public familiarity, its acceptability remains highly rated. Circumvention, on the other hand, is a concern, as facial recognition is vulnerable to spoofing attacks, but ongoing research in Presentation Attack Detection (PAD) aims to mitigate this. 

\textbf{Fingerprint} biometrics, similarly to face, has exhibited high ratings across most properties except circumvention, rated medium due to concerns over presentation attacks, spoofing, and adversarial manipulation. Experts emphasized the extreme rarity of identical fingerprints, supporting the high uniqueness of the modality. The collectability of samples has improved with modern sensors, even though environmental factors still affect the quality. Overall, fingerprints remain highly reliable biometric traits, with ongoing advancements in capturing and processing.

\textbf{Gait} biometrics received high ratings for acceptability, circumvention, collectability, and universality, with medium ratings for other properties. Experts praised its ease of capture and difficulty of imitation, regardless of its uniqueness, which is debatable since it is sensitive to external factors such as terrain, footwear, and backpacks. Compared to earlier studies, gait ratings have improved across most properties, particularly performance, which is considered one of the lowest among other biometrics due to challenges such as walking speed and external conditions. There is debate over the acceptability and circumvention of gait biometrics, while some experts value its low privacy, others doubt its reliability for security applications.

\textbf{Hand Geometry} biometrics consistently received high ratings for most properties except circumvention and performance, which were rated medium. Experts supported its ease of acquisition and general user acceptance but highlighted limited distinctiveness compared to more discriminative modalities like fingerprint. Even though it is rated medium for circumvention, it remains a concern as it is susceptible to presentation attacks. Despite its strong practical usability, hand geometry's limited uniqueness reduces its applicability in high-security contexts.

\textbf{Hand Vein} biometrics achieved high to very high ratings across all properties. Experts emphasized the modality stability and difficulty in replicating, observing improvements over previous evaluations. Circumvention rated as 4.5 indicates high resistance to vulnerabilities due to the modality's robustness. Although there were no expert comments on this modality, the overall results indicate promising reliability for future biometric applications.

\textbf{Iris} biometrics received high to very high ratings for most properties, including universality, uniqueness, permanence, and performance, with a medium rating for collectability. Experts noted the iris's strong distinctiveness and protected nature, though capturing high-quality images requires user cooperation and specialized devices. Acceptability was lower than expected due to data capture inconvenience. Experts' opinions vary when it comes to circumvention; some found it hard to bypass, while others highlight concerns over lenses and other presentation attacks. 

\textbf{Keystroke} biometrics were rated high for all existing properties, especially universality. Experts pointed out the accessibility to typing data and its high potential for usage in continuous authentication, although typing behavior may change over time, affecting the permeance rating. Similarly, experts recognized the circumvention risks of using keystroke biometrics but argued that they are manageable, as accurate imitation of typing dynamics remains difficult.

\textbf{Retinal scan} biometrics received very high ratings across most properties. Significant differences were observed in collectability and circumvention ratings compared to previous studies. This highlights the modality's suitability for secure biometric applications, although a potential bias may be observed due to the limited number of expert evaluations. Therefore, the absence of expert justification for this biometric limits further validation.

\textbf{Signature} biometrics achieved high ratings in universality, collectability, acceptability, and performance, with a low rating given to circumvention and a medium given to permanence. Compared to earlier assessments, signature biometric performance has improved significantly after the introduction of dynamic signature analysis. Even though signatures are widely acceptable, user-friendly, and easy to collect, they remain vulnerable to skilled forgery, which can compromise the uniqueness of the modality and its applicability to secure applications.

\textbf{Voice} biometrics showed varied ratings across properties. Where performance, uniqueness, permanence, and universality were rated high. While circumvention scored low due to the ease of voice imitation. Accessibility and collectability were rated medium, as voice data can be obtained from users despite privacy concerns; however, these processes are affected by aging and environmental factors. Expert opinions on circumvention varied, with some noting that voice recognition systems can be bypassed through physical or digital attacks, while others argued that spoofing remains difficult due to the voice’s high uniqueness.

\begin{table}[h!]
\footnotesize
\renewcommand{\arraystretch}{0.99}
\small
\setlength{\tabcolsep}{1.9pt}
    \centering
    \footnotesize
    \caption{ \textbf{Comparison Between Subjective Expert Ratings and Recognition Performance -} \textcolor{black}{Performance of state-of-the-art biometric systems is shown on 55 datasets across 7 modalities, including the BioQuake $(\delta)$ uncertainty. 
    For each modality, the subjective expert rating for the performance property is shown.
    Legend: Uncertainty is color-coded as \colorbox{highlightgreen!50}{Optimal}, \colorbox{highlightblue!50}{Excellent}, \colorbox{highlightper!50}{Very Good}, \colorbox{highlightyellow!50}{Good}, \colorbox{highlightorange!50}{Fair}, \colorbox{highlightbrown!50}{Poor}, and \colorbox{highlightred!50}{Unnacceptable}}. }
\label{tab:ratings_connection}
\begin{tabular}{llcccccc}
\toprule
& Dataset & \thead{FNMR \\[1pt] [\%]} & $\delta_{FNMR}$ & \thead{FMR \\[1pt] [\%]} & $\delta_{FMR}$ & \thead{Expert\\Performance\\Rating} \\
\midrule

\parbox[t]{2mm}{\multirow{13}{*}{\rotatebox[origin=c]{90}{ECG}}}

& Private~\cite{kang2016ecg} & 1.90 & \cellcolor{highlightbrown!50}{0.9398} & 5.20 & \cellcolor{highlightyellow!50}{0.13067} & \multirow{13}{*}{\thead{Low \\ (2.4)}}\\

& ECG-ID ~\cite{lugovaya2005biometric} & 2.00 & \cellcolor{highlightper!50}{0.0644} & 2.00 & \cellcolor{highlightgreen!50}{0.0068}  \\
& E-HOL~\cite{moody2001impact} & 2.15 & \cellcolor{highlightper!50}{0.0614} & 2.15 & \cellcolor{highlightgreen!50}{0.0064}  \\
& MIT-BIH~\cite{moody2001impact} & 4.74 & \cellcolor{highlightper!50}{0.0568} & 4.74 & \cellcolor{highlightgreen!50}{0.0087} \\
& PTB ~\cite{bousseljot1995nutzung} & 0.59 & \cellcolor{highlightyellow!50}{0.1531} & 0.59 & \cellcolor{highlightblue!50}{0.0223} \\
& CYBHi ~\cite{da2014check} & 6.98 & \cellcolor{highlightbrown!50}{0.7959} & 6.98 & \cellcolor{highlightorange!50}{0.3638}  \\
& CYBHi ~\cite{da2014check} & 5.44 & \cellcolor{highlightbrown!50}{0.5349} & 5.44 & \cellcolor{highlightyellow!50}{0.2626} \\
& ECG-ID ~\cite{lugovaya2005biometric} & 1.52 & \cellcolor{highlightbrown!50}{0.7392} & 1.52 & \cellcolor{highlightorange!50}{0.4048} \\
& ECG-ID ~\cite{lugovaya2005biometric} & 0.26 & \cellcolor{highlightred!50}{$>1$} & 0.26 & \cellcolor{highlightred!50}{$>1$} \\
& In-house ~\cite{melzi2023ecg} & 1.28 & \cellcolor{highlightyellow!50}{0.1361} & 1.28 & \cellcolor{highlightper!50}{0.0614} \\
& In-house ~\cite{melzi2023ecg} & 1.97 & \cellcolor{highlightyellow!50}{0.1841} & 1.97 & \cellcolor{highlightper!50}{0.0856} \\
& PTB ~\cite{bousseljot1995nutzung} & 0.14 & \cellcolor{highlightred!50}{$>1$} & 0.14 & \cellcolor{highlightred!50}{$>1$}  \\
& PTB ~\cite{bousseljot1995nutzung} & 2.06 & \cellcolor{highlightred!50}{$>1$} & 2.06 & \cellcolor{highlightbrown!50}{0.5155} \\
\hline
\parbox[t]{2mm}{\multirow{5}{*}{\rotatebox[origin=c]{90}{EEG}}}& Private~\cite{maiorana2021learning} & 4.80 & \cellcolor{highlightgreen!50}{0.0021} & 4.80 & \cellcolor{highlightgreen!50}{0.0005} & \multirow{5}{*}{\thead{Medium \\ (2.9)}}\\
& EEG Motor~\cite{goldberger2000physiobank} & 1.96 & \cellcolor{highlightgreen!50}{0.0039} & 1.96 & \cellcolor{highlightgreen!50}{0.0006} \\
& bi2015a~\cite{korczowski2019brain} & 0.21 & \cellcolor{highlightorange!50}{0.3813} & 1.00 & \cellcolor{highlightper!50}{0.0930} \\
& Private~\cite{arias2023performance} & 8.50 & \cellcolor{highlightyellow!50}{0.2205} & 8.50 & \cellcolor{highlightblue!50}{0.0333} \\
& Private~\cite{tajdini2023brainwave} & 0.52 & \cellcolor{highlightorange!50}{0.4814} & 0.52 & \cellcolor{highlightper!50}{0.0748} \\
\hline
\parbox[t]{2mm}{\multirow{15}{*}{\rotatebox[origin=c]{90}{Face}}}& LFW ~\cite{huang2008labeled} & 0.37 & \cellcolor{highlightbrown!50}{0.5405} & 0.37 & \cellcolor{highlightbrown!50}{0.5405} & \multirow{15}{*}{\thead{Very High \\ (4.3)}} \\
& LEO\_LS~\cite{best2017longitudinal} & 0.34 & \cellcolor{highlightyellow!50}{0.1979} & 0.10 & \cellcolor{highlightblue!50}{0.0263} \\
& PCSO\_LS~\cite{best2017longitudinal} & 2.17 & \cellcolor{highlightblue!50}{0.0365} & 0.10 & \cellcolor{highlightgreen!50}{0.0058} \\
& IJB-B~\cite{whitelam2017iarpa} & 5.80 & \cellcolor{highlightper!50}{0.0775} & 0.01 & \cellcolor{highlightper!50}{0.0687} \\
& IJB-C~\cite{maze2018iarpa} & 4.40 & \cellcolor{highlightper!50}{0.0652} & 0.01 & \cellcolor{highlightblue!50}{0.0493} \\
& Trillion P.~\cite{deng2019arcface} & 18.44 & \cellcolor{highlightgreen!50}{0.0012} & $10^{-7}$ & \cellcolor{highlightyellow!50}{0.1060} \\
& RFW~\cite{wang2019racial} & 4.81 & \cellcolor{highlightper!50}{0.0735} & 4.81 & \cellcolor{highlightgreen!50}{0.0012} \\
& CALFW~\cite{zheng2017cross} & 7.23 & \cellcolor{highlightyellow!50}{0.1245} & 7.23 & \cellcolor{highlightyellow!50}{0.1245} \\
& IJB-B~\cite{whitelam2017iarpa} & 3.97 & \cellcolor{highlightper!50}{0.0957} & 3.97 & \cellcolor{highlightgreen!50}{0.0034} \\
& IJB-C~\cite{maze2018iarpa} & 2.41 & \cellcolor{highlightper!50}{0.0883} & 2.41 & \cellcolor{highlightgreen!50}{0.0031} \\
& LFW ~\cite{huang2008labeled} & 0.20 & \cellcolor{highlightbrown!50}{0.6666} & 0.20 & \cellcolor{highlightbrown!50}{0.6666} \\
& AgeDB~\cite{moschoglou2017agedb} & 1.50 & \cellcolor{highlightyellow!50}{0.2888} & 1.50 & \cellcolor{highlightyellow!50}{0.2888} \\
& LFW (full) ~\cite{huang2008labeled} & 0.04 & \cellcolor{highlightyellow!50}{0.1979} & 0.04 & \cellcolor{highlightblue!50}{0.0104} \\
& ColorFERET (full) \cite{phillips2000feret} & 1.83 & \cellcolor{highlightblue!50}{0.0463} & 1.83 & \cellcolor{highlightgreen!50}{0.0018} \\
& Adience (full) ~\cite{eidinger2014age} & 2.21 & \cellcolor{highlightblue!50}{0.0157} & 2.21 & \cellcolor{highlightgreen!50}{0.0002} \\
\hline
\parbox[t]{2mm}{\multirow{6}{*}{\rotatebox[origin=c]{90}{Fingerprint}}}& FVC2002~\cite{maio2002fvc2002} & 1.57 & \cellcolor{highlightyellow!50}{0.2729} & 1.57 & \cellcolor{highlightyellow!50}{0.2123} & \multirow{6}{*}{\thead{Very High \\ (4.6)}}\\
& FVC2002~\cite{maio2002fvc2002} & 1.93 & \cellcolor{highlightyellow!50}{0.2591} & 1.93 & \cellcolor{highlightyellow!50}{0.1956} \\
& FVC2000~\cite{maio2002fvc2000} & 4.68 & \cellcolor{highlightorange!50}{0.3561} & 0.10 & \cellcolor{highlightbrown!50}{0.6061} \\
& FVC2006~\cite{cappelli2007fingerprint} & 0.15 & \cellcolor{highlightbrown!50}{0.5051} & 0.15 & \cellcolor{highlightorange!50}{0.4796} \\
& FVC2004~\cite{maio2004fvc2004} & 2.18 & \cellcolor{highlightyellow!50}{0.2457} & 0.10 & \cellcolor{highlightbrown!50}{0.8081} \\
& LFIW~\cite{liu2024latent} & 22.82 & \cellcolor{highlightblue!50}{0.0105} & 22.82 & \cellcolor{highlightgreen!50}{0.0003} \\
\hline
\parbox[t]{2mm}{\multirow{4}{*}{\rotatebox[origin=c]{90}{Gait}}}& Private~\cite{xu2018keh} & 12.10 & \cellcolor{highlightyellow!50}{0.1653} & 12.10 & \cellcolor{highlightyellow!50}{0.1653} &
\multirow{4}{*}{\thead{Medium \\ (2.7)}}\\
& Private~\cite{xu2018keh} & 6.00 & \cellcolor{highlightyellow!50}{0.2333} & 6.00 & \cellcolor{highlightyellow!50}{0.1666} \\
& Private~\cite{zhang2020deepkey} & 0.39 & \cellcolor{highlightbrown!50}{0.7692} & 0.39 & \cellcolor{highlightbrown!50}{0.7692} \\
& Private~\cite{ji2021one} & 22.20 & \cellcolor{highlightblue!50}{0.0303} & 22.2 & \cellcolor{highlightgreen!50}{0.0096} \\
\hline
\parbox[t]{2mm}{\multirow{4}{*}{\rotatebox[origin=c]{90}{Iris}}}& CASIA-IrisV1~\cite{casiairisv1} & 0.25 & \cellcolor{highlightbrown!50}{0.6956} & 0.25 & \cellcolor{highlightyellow!50}{0.1881} & \multirow{4}{*}{\thead{Very High \\ (4.6)}}\\
& CASIA-IrisV3 ~\cite{casiairisv3} & 0.54 & \cellcolor{highlightorange!50}{0.4986} & 0.54 & \cellcolor{highlightblue!50}{0.0431} \\
& CASIA-IrisV3 ~\cite{casiairisv3} & 1.93 & \cellcolor{highlightorange!50}{0.4811} & 1.93 & \cellcolor{highlightorange!50}{0.4811}  \\
& CASIA-IrisV3 ~\cite{casiairisv3} & 2.00 & \cellcolor{highlightbrown!50}{0.625} & 0.80 & \cellcolor{highlightblue!50}{0.0307} \\
\hline
\parbox[t]{2mm}{\multirow{8}{*}{\rotatebox[origin=c]{90}{Keystroke}}}& Private~\cite{hwang2009keystroke} & 4.00 & \cellcolor{highlightorange!50}{0.3125} & 4.00 & \cellcolor{highlightyellow!50}{0.25} & \multirow{8}{*}{\thead{High \\ (4.0)}}\\
& Giot~\cite{giot2009greyc} & 6.69 & \cellcolor{highlightper!50}{0.0978} & 6.69 & \cellcolor{highlightblue!50}{0.0401} \\
& Sheng~\cite{sheng2005parallel} & 4.13 & \cellcolor{highlightyellow!50}{0.2959} & 4.13 & \cellcolor{highlightper!50}{0.0559} \\
& Private~\cite{locklear2014continuous} & 4.55 & \cellcolor{highlightblue!50}{0.0234} & 4.55 & \cellcolor{highlightgreen!50}{0.0011}  \\
& Buffalo~\cite{sun2016shared} & 2.36 & \cellcolor{highlightblue!50}{0.0247} & 2.36 & \cellcolor{highlightgreen!50}{0.002} \\
& Clarkson~\cite{murphy2017shared} & 5.97 & \cellcolor{highlightgreen!50}{0.0007} & 5.97 & \cellcolor{highlightgreen!50}{0.0068} \\
& Kim~\cite{kim2020freely} & 0.02 & \cellcolor{highlightred!50}{$>1$} & 0.02 & \cellcolor{highlightbrown!50}{0.8163} \\
& Private~\cite{stylios2021bioprivacy} & 0.02 & \cellcolor{highlightred!50}{$>1$} & 0.02 & \cellcolor{highlightred!50}{$>1$} \\
\bottomrule
\end{tabular}
\end{table}

\subsection{Expert Agreement}
\begin{figure}[t]
    \includegraphics[width=0.48\textwidth]{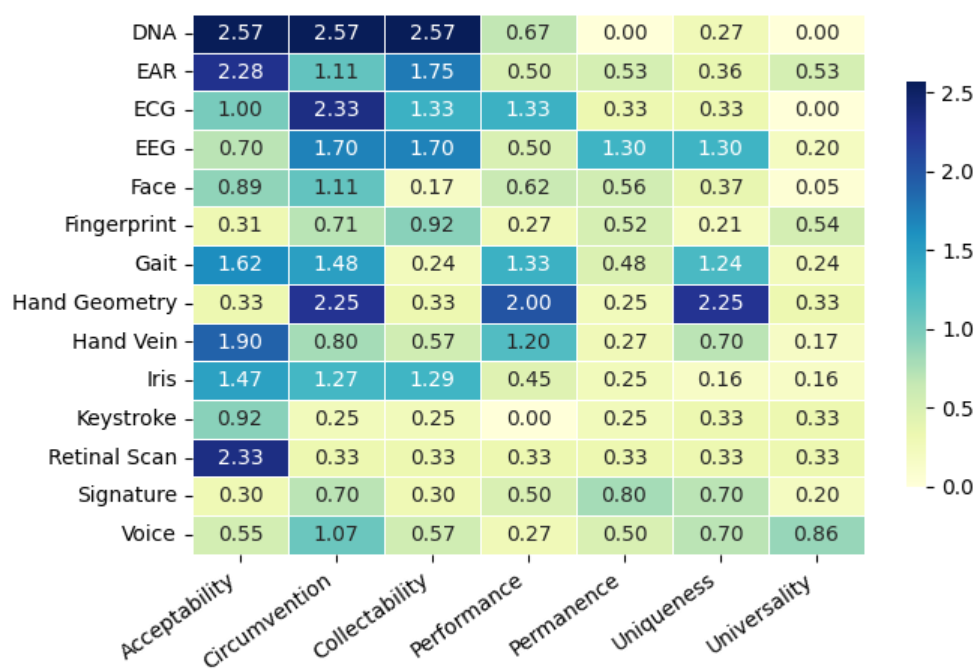}
    \caption{\textbf{Variance of Expert Rating Across Modalities -} 
    The variances between the expert ratings across biometric modalities and properties are shown. Higher variance (dark blue) indicates greater disagreement among experts, while lower variance (light yellow) indicates strong agreement. \vspace{-0.5em} 
    } 
    \label{fig:variance}
\end{figure}

To assess the level of agreement among experts regarding their ratings for each modality, the variance (standard deviation) for each modality-property pair was computed, as illustrated in Figure \ref{fig:variance}. Variance acts as a quantitative measure of consensus, where low variance indicates strong agreement and high variance signals divergence in opinions. The analysis revealed that experts showed high agreement on properties such as universality and permeance across most modalities, with particularly strong consensus for DNA, ECG and face. In comparison, significant disagreement is observed for properties like acceptability, circumvention and collectability, where modalities such as DNA and Hand Geometry have noticeably high variance scores. For instance, DNA's circumvention rating had a variance of 2.57, while hand geometry showed elevated variance for circumvention (2.25), performance (2.00) and uniqueness (2.25). Agreement on performance ratings was generally the strongest, with only few modalities displaying high variance for this property. While most experts agreed on many aspects of biometric modality properties, disagreement around circumvention and user acceptability highlight critical challenges that must be addressed, especially when considering modalities for higher-security applications.

\subsection{Investigating the Alignment Between Expert Ratings and Measurable Performance}

To investigate the alignment between expert ratings on the performance of biometric modalities and the objectively measurable performance of existing methods and datasets, including the associated uncertainty, we build upon the uncertainty analysis presented in \cite{fallahi2025reliability}. That study introduced BioQuake, a metric for quantifying the reliability of biometric benchmarks by measuring the deviation between reported and actual error rates. Specifically, BioQuake scores capture the extent to which performance metrics may be misleading due to issues such as limited or biased data. For instance, a BioQuake score exceeding a threshold of $\delta = 0.1$ implies that the reported error rate may deviate by more than 10\% from its true value.

In our work, we selected a subset of 55 biometric datasets from the original 62 analyzed in \cite{fallahi2025reliability}, corresponding to the seven biometric modalities for which expert performance ratings are available. We used this subset to compare expert perceptions on the biometric property of Performance with empirical performance and uncertainty metrics. Table \ref{tab:ratings_connection} presents this comparison, showing the current state-of-the-art performance for each dataset in terms of False Match Rate (FMR) and False Non-Match Rate (FNMR) \cite{InformationTechnologyBiometric2021}, alongside the expert-assigned performance ratings for each modality.

Comparing the performances and corresponding uncertainties to expert opinions, shows that expert ratings generally align with empirical uncertainties. Modalities such as ECG and EEG, rated medium-to-low in performance, correspond to datasets with higher uncertainty. Conversely, face biometrics rated highly by experts demonstrate lower uncertainty, benefiting from larger dataset sizes and broader evaluation coverage, except for smaller benchmarks like LFW. However, slight discrepancies were observed in some modalities, particularly keystroke dynamics, where experts indicated high performance, yet dataset-level uncertainty reflected greater variability and less consistent reliability. These observations underscore the importance of complementing expert perceptions with rigorous empirical evaluations when assessing biometric system reliability.


\section{Conclusion}
Evaluating the suitability of biometric modalities for applications, specifically authentication, requires assessing them against key properties. The widely used 1998 qualitative comparison framework no longer captures advances and challenges of current biometric systems. This paper revisits that evaluation by collecting expert ratings across 14 different modalities and combining numerical scoring and qualitative insights. The revised comparison table reveals substantial shifts in expert perception, highlighting improved confidence in face, fingerprint, and iris, and growing interest in emerging traits like EEG and keystroke dynamics. In addition, a conducted analysis on experts' opinions showed an overall strong agreement on properties such as universality and permanence, concurrently revealing greater variations in properties like circumvention and acceptability. To support this reassessment, we compared expert-rated performance to dataset-level uncertainty. While observing an overall alignment, discrepancies in some modalities underscore the importance of validating expert judgment with empirical evidence in future works. Our findings provide an updated foundation for selecting biometric modalities and suggest concrete research directions to help reduce expert disagreement in future work. Although this study focused on a research-oriented evaluation by prioritizing academic experts, future extensions could broaden the scope to include perspectives from industry and end-users.


\clearpage
\newpage
{\small
\bibliographystyle{ieee}
\bibliography{ijcb}
}

\clearpage
\newpage

\appendix
\section*{Supplementary Material}
\addcontentsline{toc}{section}{Supplementary Material}

This section provides additional material supporting the main method in the paper.

\subsection*{Survey Questionnaire}

The full set of expert survey questions is listed below, grouped by biometric modality where applicable. Questions Q2–Q43 are repeated for each of the 14 biometric modalities selected in Q1, following a consistent structure: a property assessment using a 5-point Likert scale, a justification prompt, and a confidence rating. These are shown under each modality heading for clarity. The remaining questions (Q1, Q44–Q47) are general and appear as standalone items outside of any modality grouping.

\begin{itemize}
   \item Q1: Please indicate which of the following biometric modalities you are familiar with. Check all that apply. (Here familiar refers to modalities you have worked with or have experience with)

\noindent
$\Box$ DNA \quad \quad \quad
$\Box$ Ear \quad
$\Box$ ECG \quad
$\Box$ EEG \quad
$\Box$ Face \\
$\Box$ Fingerprint \quad
$\Box$ Gait \quad
$\Box$ Hand Geometry \\
$\Box$ Hand Vein \quad
$\Box$ Iris \quad
$\Box$ Keystrokes \\
$\Box$ Retinal Scan \quad
$\Box$ Signature \quad
$\Box$ Voice

\end{itemize}

\begin{figure*}[ht]
    \centering
    \includegraphics[width=0.85\textwidth]{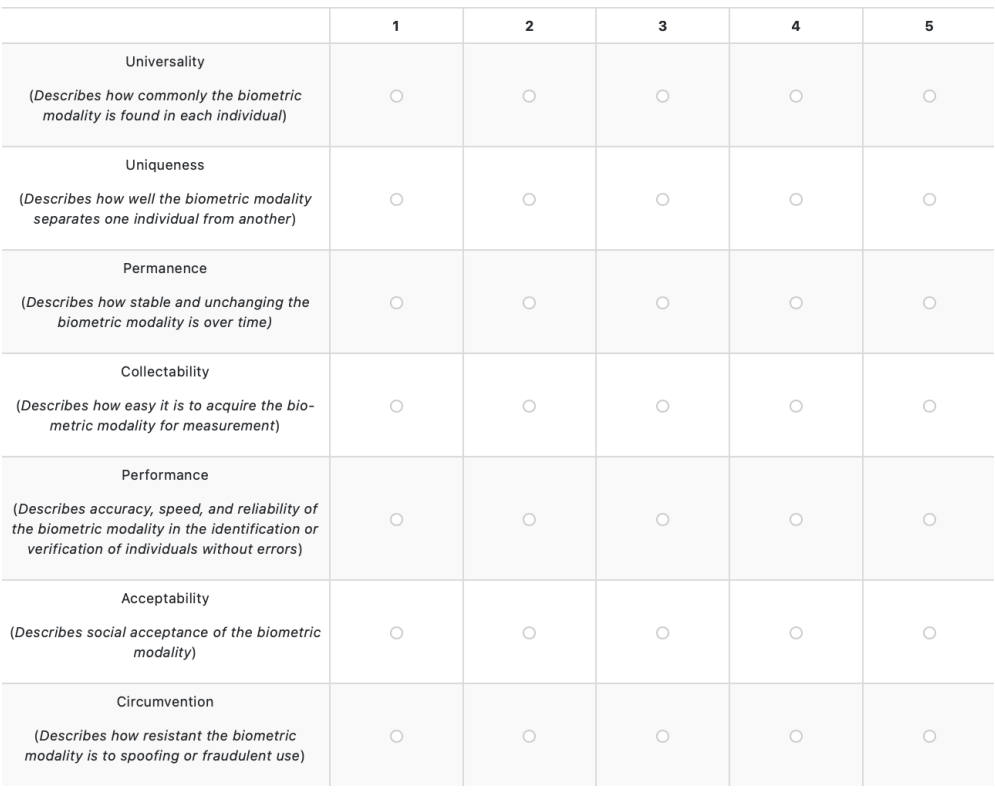}
    \caption{\textbf{Biometric Property Rating Matrix.} Experts rated each modality across seven biometric properties using a 5-point Likert scale ranging from 1 (Very Low) to 5 (Very High).}
    \label{fig:property-matrix}
\end{figure*}

\subsection*{DNA}
\begin{itemize}
    \item Q2: Please assess the extent to which \textbf{DNA} exhibits the properties shown in Figure~\ref{fig:property-matrix}, using a scale of 1 to 5, where 1 denotes very low and 5 denotes very high.
    \item Q3: Explain your rating, as shown in Figure \ref{fig:open-text}.  *(Open text input)*
    \item Q4: How confident are you about the provided ratings for properties? Answer options: Not confident, Slightly confident, Somewhat confident, Quite confident or Very confident.
\end{itemize}

\subsection*{Ear}
\begin{itemize}
    \item Q5: Please assess the extent to which \textbf{Ear} exhibits the following properties,  using a scale of 1 to 5, where 1 denotes very low and 5 denotes very high. Answer options similar to that of question Q2, Figure \ref{fig:property-matrix}. 
    \item Q6: Explain your rating, as shown in Figure \ref{fig:open-text}.  *(Open text input)*
    \item Q7: How confident are you about the provided ratings for properties? Answer options: Not confident, Slightly confident, Somewhat confident, Quite confident or Very confident.
\end{itemize}

\subsection*{ECG}
\begin{itemize}
    \item Q8: Please assess the extent to which \textbf{ECG} exhibits the following properties,  using a scale of 1 to 5, where 1 denotes very low and 5 denotes very high. Answer options similar to that of question Q2, Figure \ref{fig:property-matrix}.
    \item Q9: Explain your rating, as shown in Figure \ref{fig:open-text}. *(Open text input)*
    \item Q10: How confident are you about the provided ratings for properties? Answer options: Not confident, Slightly confident, Somewhat confident, Quite confident or Very confident.
\end{itemize}

\subsection*{EEG}
\begin{itemize}
    \item Q11: Please assess the extent to which \textbf{EEG} exhibits the following properties,  using a scale of 1 to 5, where 1 denotes very low and 5 denotes very high. Answer options similar to that of question Q2, Figure \ref{fig:property-matrix}.
    \item Q12: Explain your rating, as shown in Figure \ref{fig:open-text}. *(Open text input)*
    \item Q13: How confident are you about the provided ratings for properties? Answer options: Not confident, Slightly confident, Somewhat confident, Quite confident or Very confident.
\end{itemize}

\subsection*{Face}
\begin{itemize}
    \item Q14: Please assess the extent to which \textbf{Face} exhibits the following properties,  using a scale of 1 to 5, where 1 denotes very low and 5 denotes very high. Answer options similar to that of question Q2, Figure \ref{fig:property-matrix}.
    \item Q15: Explain your rating, as shown in Figure \ref{fig:open-text}.  *(Open text input)*
    \item Q16: How confident are you about the provided ratings for properties? Answer options: Not confident, Slightly confident, Somewhat confident, Quite confident or Very confident.
\end{itemize}

\subsection*{Fingerprint}
\begin{figure*}[ht]
    \centering
    \includegraphics[width=0.95\textwidth]{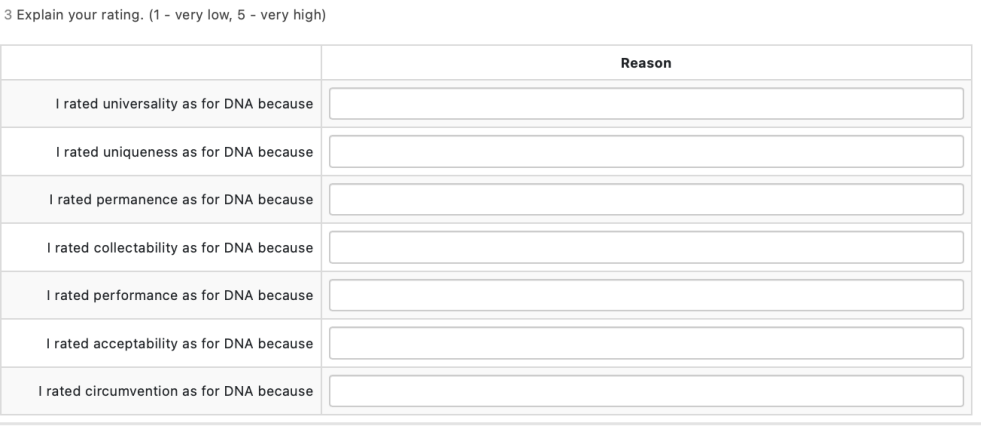}
    \caption{\textbf{Justification Entry Table for Biometric Property Ratings.} For each selected modality, experts were asked to explain their property ratings individually (Q3), providing a brief reason for each value assigned in the matrix. This format is repeated for all evaluated modalities.}
    \label{fig:open-text}
\end{figure*}

\begin{itemize}
    \item Q17: Please assess the extent to which \textbf{Fingerprint} exhibits the following properties,  using a scale of 1 to 5, where 1 denotes very low and 5 denotes very high. Answer options similar to that of question Q2, Figure \ref{fig:property-matrix}.
    \item Q18: Explain your rating, as shown in Figure \ref{fig:open-text}. *(Open text input)*
    \item Q19: How confident are you about the provided ratings for properties? Answer options: Not confident, Slightly confident, Somewhat confident, Quite confident or Very confident.
\end{itemize}

\subsection*{Gait}
\begin{itemize}
    \item Q20: Please assess the extent to which \textbf{Gait} exhibits the following properties,  using a scale of 1 to 5, where 1 denotes very low and 5 denotes very high. Answer options similar to that of question Q2, Figure \ref{fig:property-matrix}.
    \item Q21: Explain your rating, as shown in Figure \ref{fig:open-text}.  *(Open text input)*
    \item Q22: How confident are you about the provided ratings for properties? Answer options: Not confident, Slightly confident, Somewhat confident, Quite confident or Very confident.
\end{itemize}

\subsection*{Hand Geometry}
\begin{itemize}
    \item Q23: Please assess the extent to which \textbf{Hand Geometry} exhibits the following properties,  using a scale of 1 to 5, where 1 denotes very low and 5 denotes very high. Answer options similar to that of question Q2, Figure \ref{fig:property-matrix}.
    \item Q24: Explain your rating, as shown in Figure \ref{fig:open-text}.  *(Open text input)*
    \item Q25: How confident are you about the provided ratings for properties? Answer options: Not confident, Slightly confident, Somewhat confident, Quite confident or Very confident.
\end{itemize}

\subsection*{Hand Vein}
\begin{itemize}
    \item Q26: Please assess the extent to which \textbf{Hand Vein} exhibits the following properties,  using a scale of 1 to 5, where 1 denotes very low and 5 denotes very high. Answer options similar to that of question Q2, Figure \ref{fig:property-matrix}.
    \item Q27: Explain your rating, as shown in Figure \ref{fig:open-text}.  *(Open text input)*
    \item Q28: How confident are you about the provided ratings for properties? Answer options: Not confident, Slightly confident, Somewhat confident, Quite confident or Very confident.
\end{itemize}

\subsection*{Iris}
\begin{itemize}
    \item Q29: Please assess the extent to which \textbf{Iris} exhibits the following properties,  using a scale of 1 to 5, where 1 denotes very low and 5 denotes very high. Answer options similar to that of question Q2, Figure \ref{fig:property-matrix}.
    \item Q30: Explain your rating, as shown in Figure \ref{fig:open-text}. *(Open text input)*
    \item Q31: How confident are you about the provided ratings for properties? Answer options: Not confident, Slightly confident, Somewhat confident, Quite confident or Very confident.
\end{itemize}

\subsection*{Keystroke}
\begin{itemize}
    \item Q32: Please assess the extent to which \textbf{Keystroke} exhibits the following properties,  using a scale of 1 to 5, where 1 denotes very low and 5 denotes very high. Answer options similar to that of question Q2, Figure \ref{fig:property-matrix}.
    \item Q33: Explain your rating, as shown in Figure \ref{fig:open-text}.  *(Open text input)*
    \item Q34: How confident are you about the provided ratings for properties? Answer options: Not confident, Slightly confident, Somewhat confident, Quite confident or Very confident.
\end{itemize}

\subsection*{Retinal Scan}
\begin{itemize}
    \item Q35: Please assess the extent to which \textbf{Retinal Scan} exhibits the following properties,  using a scale of 1 to 5, where 1 denotes very low and 5 denotes very high. Answer options similar to that of question Q2, Figure \ref{fig:property-matrix}.
    \item Q36: Explain your rating, as shown in Figure \ref{fig:open-text}.  *(Open text input)*
    \item Q37: How confident are you about the provided ratings for properties? Answer options: Not confident, Slightly confident, Somewhat confident, Quite confident or Very confident.
\end{itemize}

\subsection*{Signature}
\begin{itemize}
    \item Q38: Please assess the extent to which \textbf{Signature} exhibits the following properties,  using a scale of 1 to 5, where 1 denotes very low and 5 denotes very high. Answer options similar to that of question Q2, Figure \ref{fig:property-matrix}.
    \item Q39: Explain your rating, as shown in Figure \ref{fig:open-text}.  *(Open text input)*
    \item Q40: How confident are you about the provided ratings for properties? Answer options: Not confident, Slightly confident, Somewhat confident, Quite confident or Very confident.
\end{itemize}

\subsection*{Voice}
\begin{itemize}
    \item Q41: Please assess the extent to which \textbf{Voice} exhibits the following properties,  using a scale of 1 to 5, where 1 denotes very low and 5 denotes very high. Answer options similar to that of question Q2, Figure \ref{fig:property-matrix}.
    \item Q42: Explain your rating, as shown in Figure \ref{fig:open-text}.  *(Open text input)*
    \item Q43: How confident are you about the provided ratings for properties? Answer options: Not confident, Slightly confident, Somewhat confident, Quite confident or Very confident.
\end{itemize}

\subsection*{General Questions}
\begin{itemize}
    \item Q44: In your opinion, are there any emerging biometric modalities that should be considered in such assessments? If yes, please describe them.
    \item Q45: Do you have any additional comments or suggestions for improving the assessment of biometric modalities?
    \item Q46: What is your age? Answer options: 21-30 years old, 31-40 years old, 41-50 years old, 51-60 years old, 61-70 years old, 71+ years old, Prefer not to say or No answer.
    \item Q47: Where do you currently reside? North America, Europe, Africa South, America, Asia, Australia, Caribbean Islands, Pacific Islands, Other, Prefer not to say, No answer.
\end{itemize}

\end{document}